# Controllable Evidence Selection in Retrieval-Augmented Question Answering via Deterministic Utility Gating

Victor P. Unda, M.A.[i]
Independent Researcher, Applied AI Systems
Email: victorunda62@gmail.com

## Abstract

Many modern AI question-answering systems convert text into vectors and retrieve the closest matches to a user's question. This works well for finding text on the same topic. However, similarity scores alone do not explain why some retrieved text is usable as evidence while other, equally similar text is not.

When multiple candidates receive similar scores, the system may retrieve sentences that repeat the same idea, sentences that are incomplete, or sentences that state a fact but answer a different question than the one being asked. Each candidate is treated as a single evidence unit, such as one sentence from unstructured text or one record from structured data. These units may be well-formed and factually correct, but similarity alone does not determine whether they can support the specific question or task.

Prior RAG work describes limits of naive retrieval pipelines and common mitigations such as query rewriting, reranking, and context compression (Gao et al., 2024). In the prototype presented here, the same issue appears in a concrete form: many candidates score similarly, but no single sentence independently states the condition the question depends on.

This limitation becomes most visible when a question depends on specific facts rather than semantic similarity alone. In the prototype evaluation, this occurred most often when a question turns on when something applies, who it applies to, or under what conditions it holds.

In such cases, many retrieved sentences appear equally similar because they use the same general language. They may mention the same law, concept, or subject, even though only some clearly state the specific detail required by the question. Other sentences may describe related context or different conditions that do not answer the question being asked.

The system can still generate a fluent answer by combining this material. However, it becomes difficult to explain why certain sentences were used and others were not. From an inspection or audit perspective, there is no explicit rule separating sentences that function as evidence from sentences that are merely relevant to the query.

This paper presents a deterministic alternative based on a working prototype. The framework introduces Meaning–Utility Estimation (MUE) and Diversity–Utility Estimation (DUE), fixed scoring and redundancy-control procedures that determine evidence admissibility before the system produces an answer. Rather than treating the most similar text as usable by default, the system explicitly evaluates whether a sentence or record can serve as evidence. In the prototype, a unit of text is used only if it clearly contains the fact, definition, rule, or condition required by the task. Units are not merged, expanded, or supplemented by nearby context. If no unit in the dataset independently contains the required information, no answer is produced.

Evidence is evaluated one sentence or record at a time using a small set of visible, deterministic signals. These signals measure semantic relatedness, explicit term coverage, dataset-level conceptual distinctiveness, and redundancy relative to already selected evidence. All scoring rules are fixed in advance and require no training or fine-tuning. Given the same data and input, the system produces the same result. In practice, this approach produces small evidence sets in which each selected unit independently satisfies the task requirements. Redundant material is suppressed, and sentences that do not clearly express the required condition or rule are excluded.

As a result, every output corresponds directly to specific sentences in the dataset, and the absence of an output corresponds to the absence of such sentences. Before any answer is produced, the system applies an explicit evidence gate. The gate passes only when at least one evidence unit independently states the information required by the task. When the gate fails, the system returns no answer.

**Introduction**

AI systems that answer questions typically operate in two stages. First, the system searches its data and retrieves text fragments related to the user's question. Second, the system generates an answer using only those retrieved fragments. The first stage is critical because it determines what the system is allowed to treat as evidence. If the retrieved text is weak or poorly justified, the final answer may still sound reasonable but will not be clearly supported by the data.

Most modern retrieval-augmented generation (RAG) systems perform retrieval using vector embeddings. Both the user's question and stored text are converted into numerical representations, and individual sentences or records are ranked using cosine similarity. This approach is efficient and effective at identifying text that is broadly related to a question. However, similarity alone does not determine whether a piece of text is suitable to serve as evidence. Sentences may score highly because they share general meaning with the query even when they do not state the information required to answer it. As a result, retrieval often returns text that is incomplete, overly general, repeated in different wording, or topically related but not directly responsive to the question (Gao et al., 2024).

Empirical evaluations reflect this limitation. Using RAGAS, an evaluation framework for retrieval-augmented systems, Es et al. (2023) show that a system can produce a fluent and convincing response even when the retrieved text does not actually state the information needed to answer the question. In these cases, the answer may sound correct, but the supporting evidence is weak or difficult to justify. This demonstrates that retrieving text that appears related is not the same as retrieving text that can support an answer.

This limitation matters because evidence selection determines what the system is permitted to rely on. When evidence selection is implicit and driven only by similarity ranking, it becomes difficult to explain why certain sentences were chosen over others or to audit how an answer was produced. A system may generate a smooth response even when the selected text does not clearly support it.

The purpose of this work is to make the evidence-selection step explicit and controllable. Rather than treating similarity as a proxy for usefulness, I treat evidence selection as a utility problem. The central question is not which text is closest to the query, but which text is most useful as supporting evidence. To address this, I introduce Meaning–Utility Estimation (MUE) and Diversity–Utility Estimation (DUE), a deterministic scoring framework that evaluates individual sentences or records based on their usefulness for answering a specific question before any answer is generated.

The proposed framework retains a standard embedding-based retrieval backbone (SBERT + FAISS) but introduces a fixed, deterministic evidence-utility layer (MUE/DUE) and an explicit evidence gate prior to generation.

**Relation to Prior Work**

Prior retrieval-augmented generation (RAG) systems use embedding similarity to retrieve context for generation (Lewis et al., 2020; Gao et al., 2024). In many pipelines, the top retrieved passages are treated as usable context without a strict, unit-level rule for what counts as evidence. RAGAS focuses on this gap from the evaluation side by measuring whether an answer is faithful to the retrieved context, whether it answers the question, and whether the retrieved context is focused (Es et al., 2023). My framework takes the next step at runtime: instead of only scoring faithfulness after the fact, it uses deterministic per-unit checks and an evidence gate to decide whether the system is allowed to answer at all.

**Motivating Example**

Consider a simple question such as, "Do you have information about human rights?" The system operates over a collection of legal text, policy documents, historical material, and general commentary. To respond, the system first retrieves sentences that are broadly related to the question based on overall meaning, without considering whether those sentences state information that can be used as evidence.

This retrieval step returns many sentences related to the topic of human rights. These may include references to the rights of citizens, general statements about equality, constitutional language, or philosophical discussions of rights. At a topical level, these sentences appear relevant, and none are obviously incorrect. However, only some of them clearly state the information needed to answer the question.

Using similarity alone, a system can combine these sentences and produce a fluent description of human rights. The problem is not that the answer sounds wrong, but that the supporting text is difficult to justify as evidence. Some sentences restate the same idea using different wording. Others mention related concepts without stating any definition or claim. Some provide background rather than information that directly answers the question.

I refer to this outcome as a messy evidence set. The issue is not the presence of false information, but the lack of a clear rule explaining why certain sentences were used and others were not. From an inspection or governance perspective, it is unclear which sentences are doing the work of answering the question and which are merely related to the topic.

The framework presented in this paper addresses this problem directly. Instead of treating all retrieved sentences as usable evidence, each candidate sentence or record is evaluated for utility. Meaning–Utility Estimation determines whether a unit, by itself, states information that can support the question. Diversity–Utility Estimation then controls how evidence is accumulated by limiting repetition and allowing additional sentences only when they add new support.

The result is a smaller and more inspectable evidence set. The system answers only when the selected text clearly states the information required by the question, and it abstains when such statements are absent.

Contributions

This paper makes the following contributions:

1. It introduces **Meaning–Utility Estimation (MUE)**, a deterministic utility-based scoring framework for evaluating whether individual evidence units can support a query.
2. It introduces **Diversity–Utility Estimation (DUE)**, an iterative selection method that controls redundancy and ensures that additional evidence units contribute non-duplicative support.
3. It proposes an **explicit evidence gate** that permits answer generation only when predefined support conditions are satisfied.
4. It demonstrates, through a working prototype, that similarity-based retrieval alone is insufficient for reliable evidence selection in retrieval-augmented systems.

Together, these components form an evidence-first retrieval framework designed for traceability, deterministic behavior, and abstention when evidence is insufficient.

**Illustrative System Behavior**

To see why similarity alone is not sufficient, consider how a typical retrieval system behaves in practice. When a broad question such as "Do you have information about human rights?" is asked, the system retrieves sentences that share semantic similarity with the query. These sentences may mention rights, equality, legal terms, or historical discussions simply because they contain overlapping language.

Using this retrieved material, the system can produce a fluent answer describing human rights. The issue is not that the answer sounds incorrect. Rather, the supporting text does not clearly function as evidence. Some retrieved sentences restate the same idea using different wording. Others mention related concepts without stating a definition or clear claim. Some provide background information rather than information that directly answers the question.

Because these sentences appear similarly relevant at a topical level, the system has no explicit basis for preferring one over another. It cannot clearly explain why a particular sentence was selected, which sentences are essential to the answer, or which could be removed without changing it. As a result, the evidence set becomes a mixture of repetition, background material, and partial statements.

Similarity is effective for locating text that is about the same topic as a question, but it does not determine whether a sentence is suitable to serve as evidence. It cannot distinguish between sentences that directly answer the question and those that merely discuss related ideas. This makes fluent answer generation possible while leaving the evidentiary basis unclear.

**Evidence Units and Retrieval Scope**

**Definition (Evidence unit).** An evidence unit is the smallest independently evaluable element of the dataset that may be used as support for an answer. In unstructured text, an evidence unit is a single sentence. In structured or semi-structured data, an evidence unit is a single record (e.g., one row). All scoring and selection procedures in this framework operate strictly at the unit level: units are retrieved as candidates, scored individually, and selected without merging, paraphrasing, or combining multiple units into a new statement.

The system operates exclusively on evidence units, which represent the smallest pieces of text permitted to support an answer. Each unit must stand on its own. In practice, this corresponds to a single sentence from unstructured text or a single record from structured data.

Each evidence unit is preserved exactly as it appears in the source data. The system does not rewrite, merge, or paraphrase text during evidence selection. When a unit is selected, its provenance is retained, enabling direct inspection of which source statements support the final output.

Evidence units are evaluated individually. Each candidate sentence or record is tested independently to determine whether it clearly states information relevant to the question. A unit is either accepted as evidence or rejected; proximity to other relevant text does not confer evidentiary status.

Retrieval and evidence selection serve distinct roles. Retrieval is used only to collect candidates. Using similarity-based search, the system gathers sentences or records that are broadly related to the question. This stage is intentionally permissive in order to minimize missed recall.

At this point, retrieved items are treated strictly as candidates, not as evidence. Because similarity alone does not establish evidentiary sufficiency, candidates are passed to a separate utility evaluation stage.

Evidence selection is then performed using explicit utility-based rules. Meaning–Utility Estimation evaluates each candidate independently for evidentiary adequacy. Diversity–Utility Estimation controls accumulation by reducing redundancy and limiting overlap among selected units. Only candidates that satisfy these criteria are included in the final evidence set. An answer is generated only when the resulting evidence set contains sufficient explicit support. Otherwise, no answer is produced.

**Formalization.** Let $q$ denote the user query and let $U = \{u_1, \ldots, u_n\}$ denote the set of retrieved candidate evidence units. Each unit $u_i$ is evaluated independently.

Meaning–Utility Estimation assigns each unit a deterministic utility score

$$\text{MUE}(u_i, q) \in [0,1],$$

where higher values indicate stronger evidentiary usefulness. A unit is considered admissible only if it satisfies the predefined utility and anchoring conditions specified by the Evidence Gate.

Diversity–Utility Estimation then constructs the final evidence set

$$E^* \subseteq U, E^* = \text{DUE}(\{u_i \mid u_i \text{ is admissible}\}),$$

where DUE suppresses redundancy and retains only non-overlapping supporting units.

An answer is produced only if $E^* \neq \emptyset$; otherwise, no answer is produced

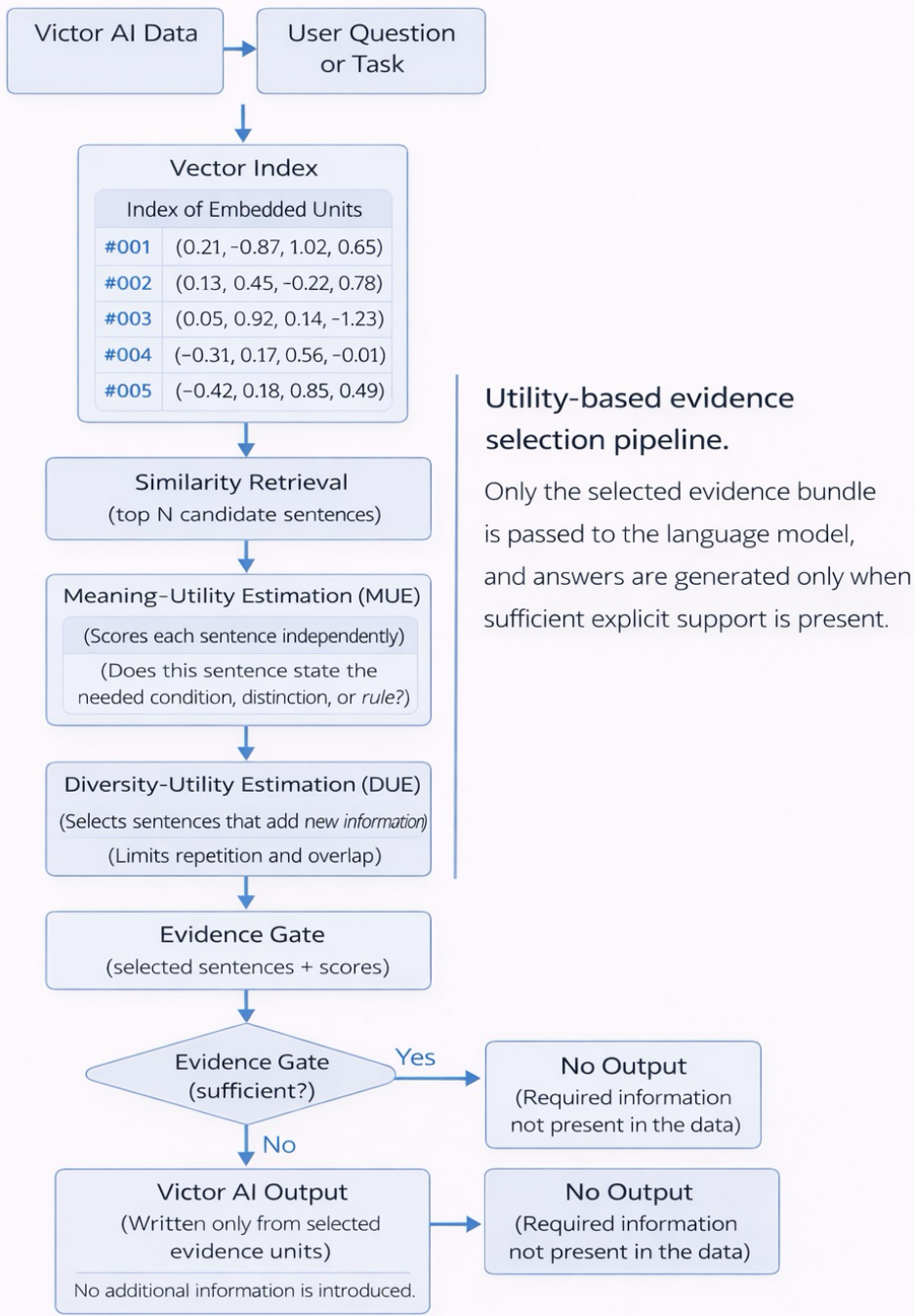

*Figure 1: Evidence-first retrieval pipeline. Similarity is used only to retrieve candidates. MUE and DUE control what may be used as evidence before any answer is generated.*

**Limitations of Similarity-Based Retrieval**

Many modern retrieval-augmented question-answering systems follow a similar retrieve-then-read pattern, although terminology varies across the literature. In general, source text is divided into chunks or passages, encoded into vector representations, and retrieved using similarity search (Lewis et al., 2020; Gao et al., 2024).

This process is effective for finding text that discusses the same general subject as the question. For example, if a question mentions "rights," the system will tend to retrieve sentences that also mention rights, citizens, laws, or related concepts. At this stage, the objective is not to determine which text answers the question, but to collect material that appears broadly relevant for further evaluation.

However, topical similarity does not guarantee evidentiary usefulness. A sentence may appear close to the query because it shares vocabulary or context, even when it does not state the information being requested. Consequently, many candidates may appear similarly relevant, even though only a subset clearly contains the definition, rule, or fact required to answer the question.

In practice, similarity-based retrieval often returns clusters of near-duplicate or weakly informative sentences. Some restate the same idea using different wording. Others mention related terms without expressing a concrete definition, rule, or factual condition. Still others provide background commentary rather than information that directly answers the query. Similarity scores alone provide no principled mechanism for separating these cases.

This limitation becomes more pronounced in structured and semi-structured data. In such settings, whether a sentence or record answers a question often depends on a specific attribute—such as a date, role, condition, or boundary between records. Similarity scoring reflects general semantic proximity but does not reliably preserve these fine-grained distinctions.

As a result, two sentences may appear highly similar even when only one satisfies the conditions implied by the query. Many questions hinge on when something applies, to whom it applies, or under what constraints it holds—distinctions that similarity scoring is not designed to enforce.

For this reason, in the proposed framework similarity-based retrieval is used only to collect candidate units. It does not determine evidentiary status. A separate utility-based stage applies explicit rules to decide which units qualify as evidence, based on whether they clearly state the required information and whether they contribute non-redundant support.

**Conceptual Importance Within the Dataset**

When the system evaluates a sentence or record, it does not treat the text as an undifferentiated block. Instead, it analyzes the individual terms that compose the unit. Terms such as "human," "rights," "citizens," or "law" are examined because both the user query and the stored corpus are processed at the lexical level.

During the initial retrieval stage, candidate units are selected based on semantic similarity to the query. Units that share vocabulary or discuss related concepts tend to be retrieved together. This behavior is effective for topical recall, but it does not determine whether a unit explicitly states information that can answer the question.

A unit may contain important keywords and still fail to express a definition, rule, or condition. To address this limitation, the framework evaluates how informative the unit's terms are within the broader dataset. Some terms appear frequently across many units and therefore provide limited discriminative value. Other terms occur in more specific contexts and better distinguish substantive statements from generic language.

Each evidence unit is therefore assigned a conceptual distinctiveness score based on corpus-level term statistics. Terms that appear widely across the dataset contribute less weight, while terms that are more distinctive contribute more. The resulting conceptual signal is normalized to a bounded range to maintain stability across datasets and to prevent common or boilerplate language from dominating the utility score. Importantly, conceptual distinctiveness reflects how informative a unit is within the corpus, rather than how often particular words repeat within the unit itself.

This distinction allows the system to separate sentences that merely mention relevant terminology from those that express concrete, dataset-consistent information. In the human-rights example, many retrieved sentences contain the word "rights," but only a subset consistently describes specific legal categories or protections. Those units receive higher conceptual distinctiveness and are treated as stronger evidence, while brief or generic mentions contribute little.

Similarity-based retrieval serves only as a candidate-generation step. Retrieved units are not treated as evidence by default. Conceptual scoring is applied afterward as part of utility evaluation to determine which candidates meaningfully support the query. A unit is not selected merely because it contains a relevant term; it is selected because it clearly states the information required by the question.

**Meaning–Utility Estimation (MUE)**

After candidate evidence units have been retrieved, the system must determine which of them are permitted to serve as evidence. Similarity scores alone do not make this determination: many units may appear equally related to a query while differing in whether they explicitly state the information required to answer it. Meaning–Utility Estimation (MUE) makes this decision explicit and rule-based.

MUE applies a direct test to each candidate evidence unit. Rather than asking which text is closest to the query, MUE evaluates whether a unit is usable as evidence for that query. Each candidate is evaluated independently, prior to evidence selection and prior to answer generation. MUE does not attempt to determine whether an answer is correct and does not judge factual truth. Instead, it asks a narrower question: does this unit, by itself, clearly state information that responds to the query? Units that satisfy this criterion are retained as admissible evidence; units that do not are excluded.

**Formal Specification**

Let $Q$ denote a user query and let
$U = \{u_1, u_2, \ldots, u_n\}$ denote the set of retrieved candidate evidence units.

For each unit $u_i$, the system computes three deterministic signals:

- **Sim$(u_i, Q)$**: semantic similarity between $u_i$ and $Q$, computed via normalized sentence embeddings and cosine similarity
- **Rel$(u_i, Q)$**: lexical coverage of key content terms from $Q$ that appear explicitly in $u_i$
- **CI$(u_i)$**: dataset-level conceptual distinctiveness of $u_i$, computed from corpus frequency statistics and bounded to the interval [0, 1] for stability

Meaning–Utility Estimation assigns each unit a utility score using a fixed, non-learned weighted combination:

$$\text{MUE}(u_i) = \lambda \cdot CI(u_i) + \mu \cdot Sim(u_i, Q) + \nu \cdot Rel(u_i, Q)$$

where $\lambda, \mu, \nu$ are predefined configuration constants that remain fixed during operation and across experiments.

**Implementation Alignment (Prototype Mapping)**

In the prototype:

- **Sim$(u_i, Q)$** is computed as cosine similarity between L2-normalized Sentence-BERT embeddings; FAISS inner-product search over normalized vectors is used for efficient candidate retrieval.
- **Rel$(u_i, Q)$** is implemented as key-term coverage: the query is filtered into content-bearing tokens (e.g., removing stopwords and very short tokens), and the system measures how many appear explicitly in the evidence unit, with an optional small fuzzy-match component for robustness.
- **CI$(u_i)$** is implemented as a corpus-frequency distinctiveness signal (IDF-style weighting) and normalized to [0, 1] for stability.

**Diversity–Utility Estimation (DUE)**

Diversity–Utility Estimation constructs an evidence set $E \subseteq U$ using a fixed iterative procedure. At each step, candidate units are evaluated based on:

1. their utility score, and
2. their similarity to already selected units,

so that near-duplicate units are suppressed and additional units are included only when they contribute non-redundant support.

**Evidence Gate**

An explicit evidence gate determines whether the system is permitted to generate an answer. The gate returns PASS only if all predefined deterministic conditions are satisfied, including:

- a minimum evidence count:

$$|E| \geq k$$

- mean lexical relevance and mean MUE over $E$ each exceeding predefined thresholds
- the existence of at least one anchor unit $u \in E$ such that

$$Rel(u, Q) \geq \tau_{rel} \text{ and } Sim(u, Q) \geq \tau_{sim}$$

where $\tau_{rel}$ and $\tau_{sim}$ are fixed minimum thresholds for lexical relevance and semantic similarity, respectively. In the prototype, these thresholds are configured deterministically (default: $\tau_{rel} = 0.30$, $\tau_{sim} = 0.35$) and remain constant across experiments.

When phrase anchoring is enabled, the gate additionally requires that at least one selected evidence unit explicitly contain a predefined high-risk phrase matched deterministically from the query (e.g., "Fourteenth Amendment"). This mechanism is intentionally limited and inspectable and can be extended by adding additional phrases to the rule set.

The prototype uses publicly available SBERT embeddings and deterministic post-retrieval filtering; no model fine-tuning is performed.

**Pipeline:**
Query → SBERT → FAISS → MUE → DUE → Evidence Gate → LLM Answer (only if PASS)

**Evidence Utility Checks**

MUE evaluates each candidate evidence unit using three deterministic signals. Each signal addresses a different weakness of similarity-based retrieval. Together, they determine whether a unit can be used as evidence for a query.

**Semantic similarity** checks whether a unit is generally related to the query. This step ensures topical alignment. It is useful for locating relevant material, but by itself it does not determine whether a unit clearly answers the query or states the required information.

**Lexical relevance** checks whether key terms from the query appear explicitly in the unit. This matters because two units can be semantically similar while using different wording. When a query refers to a specific rule, category, or named concept, units that explicitly contain those terms provide stronger evidentiary support than units that only discuss the topic in general terms.

**Conceptual importance** evaluates whether a unit uses concepts that are distinctive within the dataset. Some terms consistently appear when specific rules, categories, or definitions are being stated. Other terms occur frequently but do not help distinguish one statement from another. Units that contain dataset-distinctive concepts are therefore treated as more informative than units relying primarily on generic language.

Conceptual importance is not driven by repetition. Repeating the same term does not continue to increase a unit's score. A concept contributes value because it is present, not because it appears multiple times. This design prevents boilerplate or repetitive phrasing from being misinterpreted as strong evidence.

## Deterministic scoring

Each utility signal evaluates a different aspect of an evidence unit that matters for evidence selection. One signal checks whether the unit is about the same subject as the query. Another checks whether the unit explicitly uses the key terms referenced in the query. A third signal (CI) measures whether the unit contains concepts that are distinctive and meaningful within the dataset, rather than relying on generic or commonly repeated language.

The three signals are combined using fixed weights defined as part of the system design. These weights are set in advance and do not change based on the data, the query, or past system behavior. They are not learned or adjusted during operation. Given the same query, the same evidence units, and the same configuration, the system always produces the same result.

Because each signal is explicit, it is possible to see why one unit was accepted as evidence and another was not by inspecting the individual signal values. Evidence selection is not hidden inside a model decision. Each outcome can be traced to specific checks applied to the unit, such as whether it contains the required terms or whether it states a clear definition or rule.

This design makes evidence selection directly inspectable. A reviewer can see which units were selected, which were rejected, and why. The system does not rely on implicit model judgment to determine what counts as evidence.

Meaning–Utility Estimation is used only to determine whether a sentence or record may be admitted as evidence for a query. It does not perform reasoning or interpretation. Its role ends before any answer is generated, ensuring that evidence selection is governed entirely by explicit, rule-based criteria.

Table 1. Components of Meaning–Utility Estimation (MUE)

| Signal | What it checks | Source | Why it is needed |
|---|---|---|---|
| Semantic similarity | Whether the evidence unit is broadly about the same topic as the question | Sentence-BERT embeddings with cosine similarity | Retrieves topically related candidates but does not by itself guarantee evidentiary usefulness |
| Lexical relevance | Whether key content terms from the question appear explicitly in the evidence unit | Deterministic token overlap between query and unit | Ensures that the evidence directly reflects the terms the question depends on |
| Conceptual importance (CI) | Whether the unit contains concepts that are distinctive within the corpus | Corpus document-frequency statistics (IDF-style) | Down-weights generic language and promotes conceptually informative units |

*MUE combines signals from embeddings, raw text, and corpus statistics to evaluate whether a sentence is suitable to be used as evidence, rather than merely related to the question.*

Table 2. Example Evidence-Gate Trace (observable outputs)
Question: "Equal protection applies to state action, not private conduct."

| Metric | Observed Value |
|---|---|
| Retrieved units (n) | 6 |
| Top similarity | 0.46(rounded) |
| Mean lexical relevance | 0.07 |
| Mean utility score | 0.53 |
| Evidence Gate | FAIL |
| System output | No answer generated |

**Example Interpretation of Gate Failure**
Although the top similarity score was moderate (≈0.46), the mean lexical relevance was low (≈0.07). The retrieved sentences were related at a general legal level but did not explicitly state the condition referenced in the question. Because no individual sentence independently contained the required information, the evidence gate failed and the system declined to answer.

**Illustrative MUE Behavior**
The effect of Meaning–Utility Estimation can be seen by comparing two similar questions.

When the question is, *"Do you have information about equal protection under the Fourteenth Amendment?"*, the system retrieves sentences related to constitutional rights and legal language. These sentences are topically related, but none clearly state what equal protection is or how it applies. Because no sentence directly answers the question, the system does not select evidence and does not produce an answer. This behavior is by design. Similarity alone is not treated as sufficient justification for admitting a sentence as evidence.

The same pattern appears with broader questions. When asked, *"What rights are protected by the Fourteenth Amendment?"*, the system retrieves sentences that mention rights, amendments, and legal terms. However, none of the retrieved sentences explicitly list or describe the rights protected by the Fourteenth Amendment. As a result, the system again determines that the available text is insufficient and declines to answer.

These examples illustrate the role of Meaning–Utility Estimation. It does not interpret the law or fill in missing information. Its function is limited to verifying whether the available text clearly states the information required by the question. When the data does not contain such statements, no answer is produced. This controlled stopping behavior is a central design goal of the framework.

**Diversity–Utility Estimation (DUE)**

A common problem with similarity-based retrieval is repetition. In legal, policy, and technical corpora, the same idea is often expressed multiple times using different wording. A rule or definition may appear in statutes, court opinions, summaries, and commentary. When retrieval relies on similarity alone, many of these sentences are returned together because they are linguistically similar. Including all of these sentences does not strengthen evidentiary support. Instead, they repeat the same information and make the evidence set harder to inspect. The issue is not that the sentences are incorrect, but that they do not contribute additional support.

Diversity–Utility Estimation (DUE) is designed to control this effect. Once one sentence has been selected as evidence, the system compares each remaining candidate against the evidence already chosen. If a candidate is highly similar to an existing evidence unit, its selection priority is reduced. Sentences that merely restate the same information are suppressed, while sentences that contribute new support remain eligible.

DUE builds on the same signals used in MUE—semantic similarity to the query, lexical relevance to key terms, and conceptual importance—but introduces an additional constraint: similarity to already selected evidence. Evidence is selected iteratively, one unit at a time. At each step, the system chooses the candidate that provides the greatest additional utility without introducing redundancy.

DUE does not perform reasoning, interpretation, or summarization. It does not determine whether an answer is correct. Its role is strictly to regulate evidence accumulation. The evidence set grows only when a newly selected unit contributes information not already represented.

During evidence selection, DUE follows three operational rules:

- **Redundancy suppression.** When multiple sentences express the same information, only one is retained. Near-duplicate sentences are treated as repetition rather than additional support.
- **Incremental support.** When a sentence introduces a new fact, category, or condition not already present in the evidence set, it may be selected even if it concerns the same general topic.
- **No compensatory aggregation.** When no individual sentence clearly supports the query, DUE does not attempt to combine partial statements. In such cases, the system may terminate evidence accumulation and abstain.

Together, these rules keep evidence sets compact and interpretable. Redundant material is minimized, weak accumulation is avoided, and units are admitted only when they provide distinct support. This behavior directly supports the framework's goal of producing answers only when the underlying data provides clear, inspectable justification.

Table 3. Diversity–Utility Estimation (DUE) Selection Rules

| Situation | DUE behavior | Why this is done |
|---|---|---|
| Multiple sentences state the same information | Keep one and suppress the rest | Repetition does not increase evidentiary support |
| A sentence adds a new fact or detail | Allow it to be selected | Evidence should expand only when new support is introduced |
| A sentence is very similar to selected evidence | Lower its score | Prevents near-duplicate units from dominating the evidence set |
| No sentence clearly answers the question | Do not select evidence | The system must not guess or combine weak partial statements |

*DUE controls how evidence accumulates by favoring new information and suppressing redundancy, without attempting reasoning or interpretation.*

**Evidence Selection and Controlled Response**
After retrieval and utility scoring, the system constructs the final evidence set permitted for response generation. This set contains only individual sentences or records that satisfy both Meaning–Utility Estimation (MUE) and Diversity–Utility Estimation (DUE). All other retrieved material is discarded and plays no role in the response.

Each selected unit is preserved exactly as it appears in the source data. The system does not rewrite, summarize, or otherwise modify evidence at this stage. For every admitted unit, the system retains provenance metadata together with the signal values that justified selection. This enables direct inspection of which statements support the final output and why they were admitted.

Only the finalized evidence set is provided to the language model. The generator has no access to any other retrieved content. This constraint ensures that the produced answer can be grounded only in units that were explicitly validated as evidence.

Before answer generation, the system applies the evidence gate. If the selected evidence does not clearly state the information required to answer the query, the system returns no answer. This decision is deterministic and does not depend on language-model confidence or decoding behavior; it depends solely on whether the evidence itself contains explicit support.

This behavior becomes visible for queries that are topically related to the dataset but lack explicit evidentiary support. For example, questions about whether the Fourteenth Amendment applies to private companies may retrieve many legally related sentences. However, if none of those sentences clearly states the governing rule, the system reports that no answer is available.

The framework is therefore designed to answer only when the available text provides clear support. When such support is not present, the system does not infer, aggregate partial statements, or rely on external knowledge. In these cases, no answer is produced.

**Role of the Language Model**
In this framework, the language model does not participate in evidence selection. Its role is limited to generating a natural-language response conditioned on the finalized evidence set.
The model is provided only the units that passed MUE/DUE and the evidence gate. It does not have access to rejected candidates and does not introduce external knowledge. If no evidence is admitted, no answer is produced.
This separation improves traceability. When an answer is produced, the exact supporting units are visible. When no answer is produced, the absence of support can be traced directly to the dataset rather than to internal model behavior.

**Observed System Behavior in Practice**

In the prototype evaluation on a curated corpus of legal and historical texts, the system exhibits stable and predictable behavior under fixed retrieval and gating parameters. In the proposed framework, similarity-based retrieval is used only to collect candidate units that are broadly relevant to the query. Utility-based scoring then determines which sentences, if any, qualify as admissible evidence.

When the dataset contains sentences that explicitly state the requested information, the system produces an answer supported by a compact evidence set. When the dataset contains only partial statements, background discussion, or indirect references, no answer is produced.

This behavior reflects the intended operating principle of the framework: an output is generated only when the corpus contains at least one unit that independently satisfies the evidentiary requirement. When such text is absent, the system returns no answer rather than attempting to synthesize a response from incomplete support.

**Quantitative Observations From the Prototype**

The prototype was evaluated over a corpus of historical and legal materials using fixed retrieval parameters (TOP_K = 6, CAND_K = 30). All scoring weights and gate thresholds remained constant across runs.

The following cases illustrate the observed pattern.

    A. Condition-Specific Questions

        **Example claim:**
        "Equal protection applies to state action, not private conduct."

        **Observed metrics:**
        Top similarity ≈ 0.46
        Evidence Gate: **FAIL**

        **Interpretation:**
        Although retrieved sentences were topically related to rights and legal doctrine, none independently stated the required condition. The system therefore abstained.

        **Observation:**
        Moderate semantic similarity alone did not trigger answer generation.

B. Entity-Specific Questions

**Example claim:**
"The Fourteenth Amendment addresses civil rights."

**Observed metrics:**
Top similarity ≈ 0.65
Evidence Gate: **FAIL**

**Interpretation:**
Retrieved sentences referenced civil rights and constitutional themes but did not explicitly assert the queried relationship. The evidence gate blocked generation.

**Observation:**
Even relatively high similarity did not override the explicit evidence requirement.

C. Definition Questions

**Example query:**
"What is human rights?"

**Observed metrics:**
Top similarity ≈ 0.69
Evidence Gate: **PASS**

**Interpretation:**
Retrieved material contained explicit definitional language. Because at least one unit independently satisfied the evidentiary criteria, the system produced an answer.

**Observation:**
When explicit support is present, the system responds deterministically and with a compact evidence set.

Across repeated tests, definition-style queries with explicit statements produced higher lexical coverage and consistently passed the evidence gate. In contrast, condition-based and entity-claim queries frequently failed because no single retrieved unit independently stated the required information, even when similarity scores were moderate.

These results were stable across runs and reflect the deterministic structure of the framework.

Although the evidence gate enforces explicit support requirements, some admitted units may still reflect strong topical statements rather than fully canonical definitions. The current design prioritizes deterministic traceability and controlled abstention over aggressive semantic filtering. Further semantic tightening of admissibility criteria remains an avenue for future work.

**To evaluate the stability of the MUE configuration, multiple fixed weight settings were tested over the same question set**

| Question | Weights (λ, μ, ν) | Retrieval % | n | Mean Rel | Mean MUE | Max Sim | Max Rel | anchor_ok | phrase_ok | Gate |
|---|---|---|---|---|---|---|---|---|---|---|
| What is human rights? | (0.50,0.30,0.20) | 70% | 6 | 0.662 | 0.704 | 0.696 | 1.000 | 1 | 1 | PASS |
| What are natural rights? | (0.50,0.30,0.20) | 84% | 6 | 0.990 | 0.786 | 0.841 | 1.000 | 1 | 1 | PASS |
| What are civil rights? | (0.50,0.30,0.20) | 81% | 6 | 0.990 | 0.765 | 0.814 | 1.000 | 1 | 1 | PASS |
| The Fourteenth Amendment addresses civil rights. | (0.50,0.30,0.20) | 65% | 6 | 0.340 | 0.625 | 0.648 | 0.409 | 1 | 0 | FAIL |
| Equal protection applies to state action, not private conduct. | (0.50,0.30,0.20) | 46% | 6 | 0.073 | 0.528 | 0.456 | 0.163 | 0 | 0 | FAIL |
| When does equal protection apply? | (0.50,0.30,0.20) | 56% | 6 | 0.337 | 0.588 | 0.557 | 0.357 | 1 | 0 | FAIL |
| What is equal protection? | (0.50,0.30,0.20) | 58% | 6 | 0.493 | 0.626 | 0.581 | 0.522 | 1 | 0 | FAIL |
| What is equal protection under the Fourteenth Amendment? | (0.50,0.30,0.20) | 54% | 6 | 0.088 | 0.539 | 0.484 | 0.221 | 0 | 0 | FAIL |
| Does equal protection apply to private conduct? | (0.50,0.30,0.20) | 46% | 6 | 0.149 | 0.534 | 0.463 | 0.217 | 0 | 0 | FAIL |
| Define equal protection. | (0.50,0.30,0.20) | 56% | 6 | 0.492 | 0.632 | 0.556 | 0.522 | 1 | 0 | FAIL |
| Equal Protection Clause definition. | (0.50,0.30,0.20) |  | 6 | 0.338 | 0.590 | 0.481 | 0.355 | 1 | 0 | FAIL |
| Under what conditions does equal protection apply? | (0.50,0.30,0.20) | 57% | 6 | 0.214 | 0.568 | 0.570 | 0.230 | 0 | 0 | FAIL |
| Define civil rights. | (0.50,0.30,0.20) | 79% | 6 | 0.990 | 0.760 | 0.793 | 1.000 | 1 | 1 | PASS |

*Table 4: Sensitivity analysis of MUE weight configurations. Across multiple weight settings, definition-style queries consistently pass the evidence gate when explicit unit-level support is present, while condition- and claim-based queries fail when no single evidence unit independently satisfies the deterministic thresholds. The FINAL configuration (λ = 0.50, μ = 0.30, ν = 0.20) was selected because it produced the most stable behavior across both positive (PASS) and abstention (FAIL) cases in the updated prototype.*

**Applied System Behavior**

The proposed Meaning–Utility Estimation (MUE) and Diversity–Utility Estimation (DUE) procedures were implemented in a working prototype operating over unstructured text (sentence-level units). The design is extensible to structured and semi-structured sources by treating each record as an evidence unit evaluated under the same deterministic rules. The system begins with a similarity-based retrieval stage that returns a fixed candidate set of sentences related to the query. This step defines the evaluation scope.

All subsequent decisions are performed strictly within this retrieved candidate set. The system does not expand the search space, incorporate external material, or rely on text outside the retrieved scope. Each candidate unit is evaluated independently using MUE and then filtered iteratively by DUE to determine evidentiary admissibility.

Units that do not explicitly state the required information are rejected. Units that substantially repeat already selected content are suppressed through diversity control. Evidence is retained only when a unit independently provides usable support for the query. If no unit satisfies the admissibility criteria, the system abstains and reports that the information is not available in the dataset.

In practice, this process yields compact and inspectable evidence sets when the corpus contains explicit supporting statements. When such statements are absent, the system produces no answer. In both cases, the outcome follows deterministically from fixed rules and can be traced directly to the evaluated evidence units.

**System Behavior and Constraints**

The system operates in a fully deterministic manner. All scoring rules and thresholds are specified a priori and remain fixed during operation. The framework does not learn from prior queries, does not adapt its decision criteria based on outcomes, and does not modify the underlying language model. Consequently, identical queries over the same dataset always produce identical results.

The system relies exclusively on information explicitly stated in the dataset. If no evidence unit independently expresses the information required by the query, no answer is produced. It does not infer unstated conclusions, extend partial meanings, or combine multiple incomplete statements to construct an implied answer. This behavior is consistently observed in cases where retrieved sentences are topically related but fail to meet the evidentiary requirement.

The objective of the framework is not to maximize answer coverage but to ensure that every produced answer is directly traceable to specific evidence units in the corpus. When such units are absent, the system terminates without generating a response.

**Conclusion**

This paper presents a utility-based approach to evidence selection centered on Meaning–Utility Estimation (MUE) and Diversity–Utility Estimation (DUE). Rather than treating the most similar retrieved text as evidence by default, the framework evaluates whether individual sentences or records can serve as direct support for a specific query.

MUE evaluates each evidence unit independently and determines whether it explicitly states information relevant to the query. DUE then governs evidence accumulation by reducing redundancy and permitting additional units only when they contribute new, explicitly supported information. Evidence selection is completed prior to answer generation and follows fixed, inspectable rules rather than implicit model behavior.

The resulting pipeline implements an evidence-first retrieval process in which every generated answer is traceable to specific source units. When the available text does not explicitly support the query, no answer is produced.

By prioritizing explicit support over broad coverage, the framework emphasizes justification, transparency, and predictable behavior. It is intended for settings in which understanding why an answer was produced—or why no answer was produced—is prioritized over maximizing response coverage.

This deterministic, evidence-first design is particularly suited to knowledge-intensive environments where traceability and controlled non-response are important design requirements.

**Limitations.** The current prototype operates at sentence-level evidence granularity and uses fixed thresholds calibrated on a limited evaluation corpus. Performance in highly noisy collections, multilingual settings, or domains with heavy paraphrasing remains untested. Future work will examine adaptive thresholding, broader domain coverage, and large-scale benchmark evaluation.

---

[i] The systems and methods described in this paper are the subject of a pending patent application.